\documentclass{article}
\usepackage{spconf,amsmath,epsfig}
\usepackage{graphicx}
\usepackage{appendix}
\let\OLDthebibliography\thebibliography
\renewcommand\thebibliography[1]{
  \OLDthebibliography{#1}
  \setlength{\parskip}{0pt}
  \setlength{\itemsep}{0pt plus 0.3ex}
}
\usepackage[colorlinks,linkcolor=blue]{hyperref}
\usepackage{multicol} 
\usepackage{makecell}
\usepackage{multirow}
\begin{document}\sloppy

\def\x{{\mathbf x}}
\def\L{{\cal L}}

\title{FENet: Focusing Enhanced Network for Lane Detection}
%

\name{Liman Wang\thanks{* Co-first author}$^{*}$, Hanyang Zhong\thanks{* Co-first author}$^{*}$}
\address{University of York; \\ ssee02131@gmail.com, hanyang.zhong@york.ac.uk\\}

\maketitle

\begin{abstract}
Inspired by human driving focus, this research pioneers networks augmented with Focusing Sampling, Partial Field of View Evaluation, Enhanced FPN architecture and Directional IoU Loss - targeted innovations addressing obstacles to precise lane detection for autonomous driving. Experiments demonstrate our Focusing Sampling strategy, emphasizing vital distant details unlike uniform approaches, significantly boosts both benchmark and practical curved/distant lane recognition accuracy essential for safety. While FENetV1 achieves state-of-the-art conventional metric performance via enhancements isolating perspective-aware contexts mimicking driver vision, FENetV2 proves most reliable on the proposed Partial Field analysis. Hence we specifically recommend V2 for practical lane navigation despite fractional degradation on standard entire-image measures. Future directions include collecting on-road data and integrating complementary dual frameworks to further breakthroughs guided by human perception principles. The Code is available at \href{https://github.com/HanyangZhong/FENet}{here}.

\end{abstract}
%
%
\section{Introduction}
\label{sec:intro}

\begin{figure}[htbp]
\centerline{\includegraphics[width=1\linewidth]{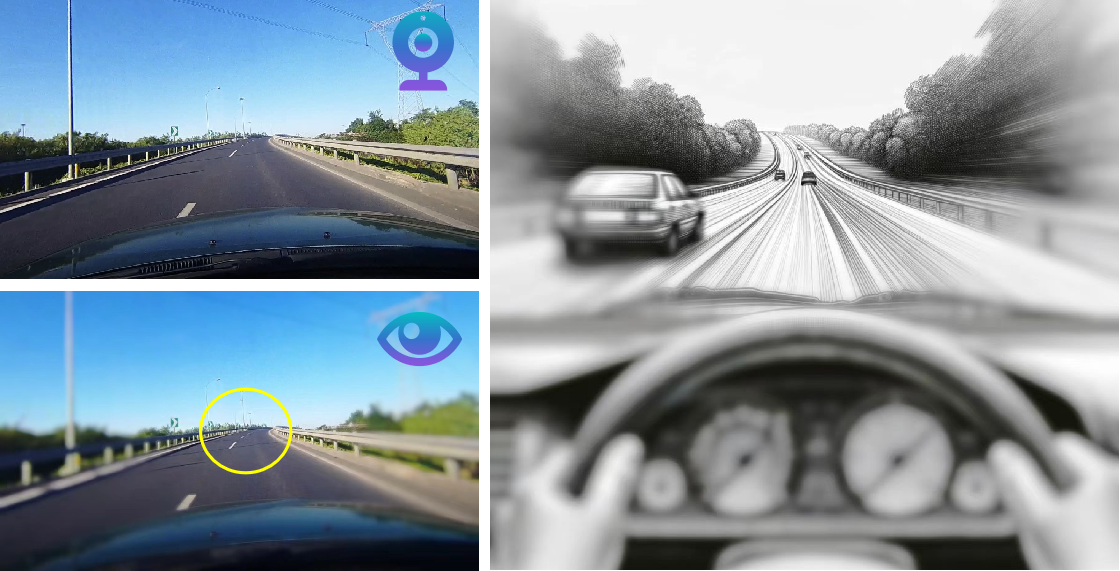}}
\caption{Skilled drivers focus their gaze far ahead on the road. The upper left image shows the \textbf{full camera view}, while the lower left image visualizes where \textbf{experienced drivers look} - far ahead along the road and lane lines.}
\label{First_fig}
\end{figure}

This study highlights the difference between human visual focus during driving and the perspectives captured by 2D cameras. Fig.~\ref{First_fig} shows that experienced drivers focus on distant road regions to anticipate path changes and steering needs, particularly around curves, looking ahead 1-2 seconds into crucial preview zones \cite{otto,lehtonen,mourant,lehtonen,boer}. Inspired by these patterns, we introduce the 'Focusing Sampling' method to better detect and regress distant lane boundaries, crucial for high-speed autonomous driving. We also propose the 'Partial Field of View Evaluation' to enhance accuracy assessments in real-world scenarios by focusing on forward road sections that align with driver focus.

\begin{figure*}[htbp]
\centering
\includegraphics[height=8cm]{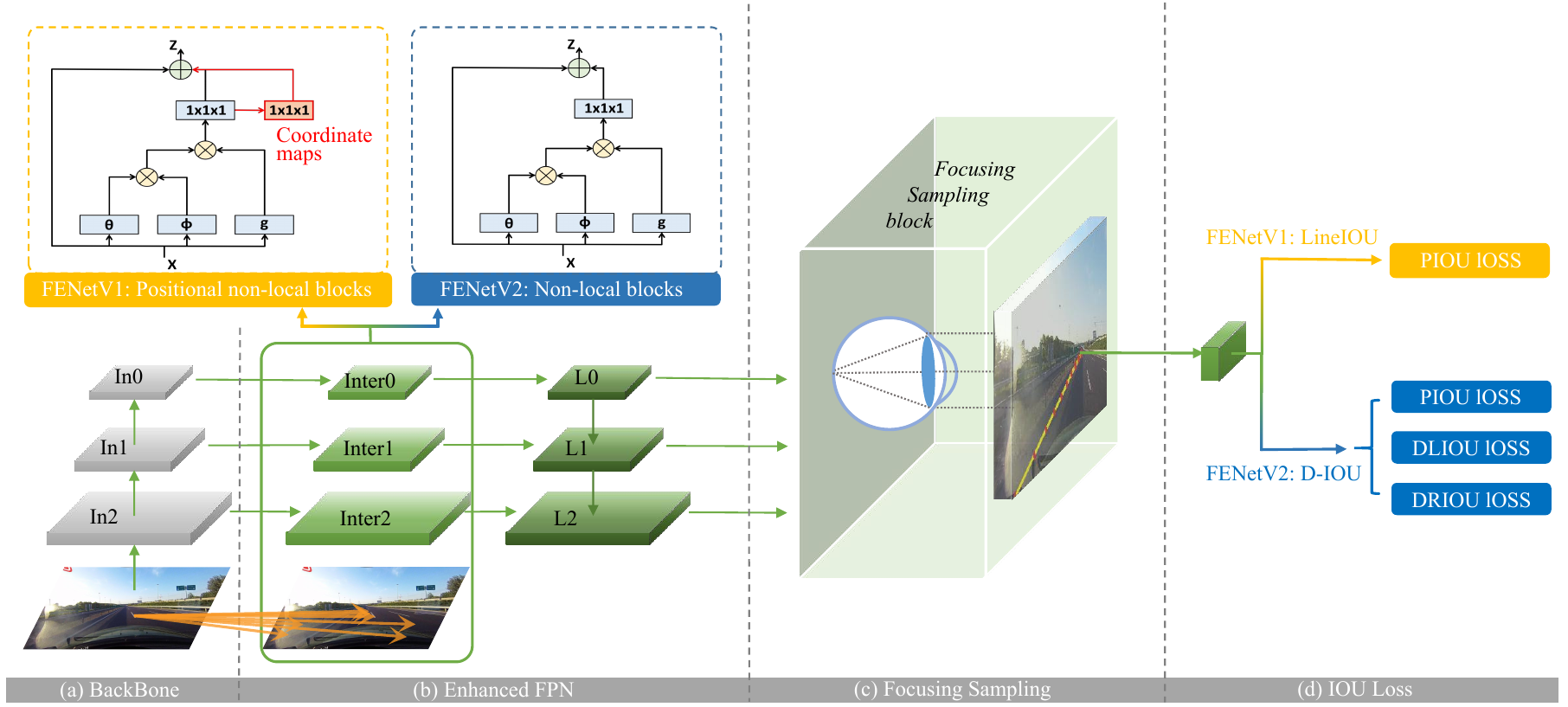}
\caption{\textbf{Architecture of the FENet lane detection framework}, which uses a DLA34 backbone and Enhanced FPN. The input layers feed into internal layers integrated with \textbf{positional non-local blocks} (for FENetV1 or {standard non-local blocks} (for FENetV2) to capture spatial context. The internal layers connect to output layers that pass through \textbf{Focusing Sampling} and either standard IOU loss (for FENetV1) or \textbf{Directional IOU loss} (for FENetV2). FENetV1 (yellow pathway) and FENetV2 (blue pathway) share the common green pathway, with distinct supplementary modules.}
\label{framework_fig}
\end{figure*}


In lane detection, widely-used models like CLRNet \cite{4} and GANet \cite{5} utilize standard multi-scale FPN architectures but fail to fully capture the necessary global spatial context for accurate road mapping crucial for safe navigation. Although transformers excel in global semantic processing, their effectiveness reduces in identifying thin, extended lane markings due to sparse cues \cite{Attention_is_all}. Lane boundaries vary greatly in scale and are influenced by different lighting and surface conditions \cite{A_sensor_fusion}. To address these issues, our study enhances the FPN by integrating either positional or standard non-local blocks, providing richer global context. Moreover, our experiments show that 'Directional IoU Loss' equals or surpasses the advantages of positional non-local blocks.


This research introduces four key innovations: (1) 'Focusing Sampling,' a training approach that emphasizes small and distant lane details; (2) 'Partial Field of View Evaluation,' new metrics for accuracy in critical forward road sections; (3) an improved FPN architecture that includes either positional or standard non-local blocks; (4) 'Directional IoU Loss,' a unique regression loss for correcting directional errors in distant lanes. FENetV1 uses positional non-local blocks for perspective-sensitive semantics, achieving top results with conventional metrics. Meanwhile, FENetV2, incorporating coordinate modeling with 'Directional IoU Loss,' excels in precisely locating distant lane boundaries. Although it may slightly lag behind FENetV1 in traditional metrics, FENetV2's focus on distant lane regression better suits real-world navigation. Overall, this work pushes for advances in regression-centric models and evaluations, focusing on techniques that enhance the depiction and assessment of critical road details for effective and safe autonomous lane detection.

\section{Related works}

Deep learning-based lane prediction methods fall into three categories: semantic segmentation, anchor-based, and parameter-based. Semantic segmentation approaches like SCNN, SAD, and Curvelanes-NAS offer pixel-level accuracy but are computationally demanding \cite{7,9,10}. Anchor-based methods, including CLRNet, GANet, and others, are quicker but struggle with accuracy in complex scenarios \cite{4,5,11,14,15,17}. Parameter-based approaches, such as those using curve modeling, prioritize efficiency but compromise accuracy \cite{20}. Our work aims to improve on these methods by addressing their accuracy and complexity limitations.


\section{Methodology}

\subsection{Focusing Sampling}

\begin{figure}[htbp]
\centerline{\includegraphics[width=1\linewidth]{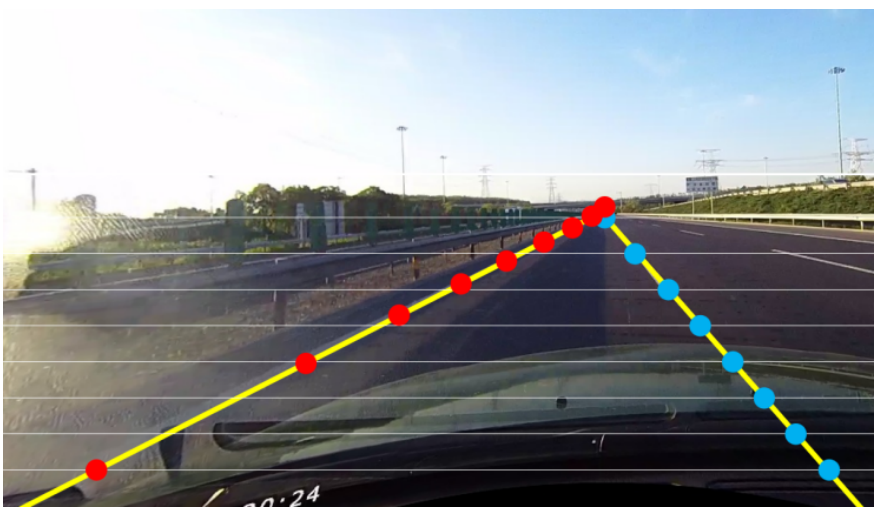}}
\caption{Visual depiction comparing \textbf{Focusing Sampling} (red dots) versus \textbf{uniform sampling} (blue dots). Focusing Sampling strategically emphasizes critical distant vanishing points along the lane while retaining informative nearby points. This accounts for perspective geometry, unlike standard uniform sampling that weights all regions equally.}
\label{Sample_fig}
\end{figure}

\textbf{Motivation.} Uniform point sampling strategies in lane detection do not prioritize visual perspective, treating all image regions equally despite the importance of distant vanishing points, especially in curves \cite{curve}. To overcome these shortcomings, we introduce Focusing Sampling, inspired by how skilled drivers focus on distant road sections to anticipate curves \cite{otto}. This method emphasizes distant details while also considering nearby points, effectively capturing the full lane geometry and addressing complex turns and curves, as shown in Fig.~\ref{Sample_fig}. Focusing Sampling aims to improve upon uniform sampling by preserving critical data and semantics essential for accurate lane prediction.

\textbf{Formulation.} Building on the foundation of uniform sampling, we propose using a logarithmic-based Focusing Sampling, for which an intuitive comparison is made in Fig.~\ref{fig7} in Appendix \ref{app_A}. The formula for this Focusing Sampling is:

{\small \begin{align}
y & = \frac{H}{N_{sample}-1 }*\log_{feature}{a_{n} }
\end{align}}


The term $a_{n} = \frac{1}{N_{sample}-1 } *i$ represents an arithmetic sequence scaling from 0 to 1 across $i$ data points, where $N_{sample}$ is the number of sample points and $H$ is the image height. To generate focusing sample points, the feature point distribution undergoes a logarithmic transformation, converting feature points into integer sample values. This method emphasizes the extraction of key semantic information. Due to the possibility of repeated values from logarithmic discretization, deduplication is applied in post-processing. The feature discretization process and its output are detailed visually in the code and further illustrated in Appendix Fig.~\ref{fig7}.

\subsection{Positional Non-local Block and Position Enhanced FPN structure}


\textbf{Motivation.} CLRNet uses the Feature Pyramid Network (FPN) architecture for lane detection \cite{4}, excelling in multi-scale feature extraction but facing challenges in detecting small targets due to its layer structure—deeper layers focus on semantics and shallower layers on fine details. To improve FPN, we incorporate non-local blocks \cite{21}, drawing on advancements from PANet and Mask R-CNN \cite{23,24}, to enhance context awareness, depth, and multi-scale capabilities. Additionally, we introduce a Position Enhanced FPN (PEFPN) module tailored for lane detection, integrating global semantics with lane coordinate modeling. This integration is detailed in Appendix \ref{app_A} Fig.~\ref{framework_fig}, showing direct infusion of position information into the FPN's multi-scale feature maps.

\textbf{Formulation.} The coordinate map equations utilize spatial indices $i$ and $j$ to index each pixel location in the feature map, where $i$ iterates vertically over the height dimension $H$, and $j$ iterates horizontally over the width dimension $W$. Specifically, $i$ ranges from 1 to $H$, indexing each row, while $j$ ranges from 1 to $W$. The x-coordinate map is calculated as follows:

{\small \begin{align}
x_{coord}(i,j) = \frac{2(j-1)}{W-1} - 1, \hspace{1mm} \text{for } i=1\ldots H, j=1\ldots W
\end{align}}

Similarly, the y-coordinate map is:

{\small \begin{align}
y_{coord}(i,j) = \frac{2(i-1)}{H-1} - 1, \hspace{1mm} \text{for } i=1\ldots H, j=1\ldots W
\end{align}}


The positional non-local block in our architecture enhances feature extraction by encoding global context and precise spatial locations. We also integrate a Focusing Sampling technique that targets specific lane segments to better track lane coordinates and directions. FENetV1 combines this with PEFPN, which incorporates direct coordinate injection and Focusing Sampling aligned with lane positions, enhancing coordinate modeling. The ablation study in Appendix \ref{app_A} demonstrates that PEFPN is a uniquely designed architecture optimized for the focused detection of lanes in FENetV1.

\subsection{Focusing Enhanced Network}


\textbf{Motivation.} In FENetV1, the PEFPN framework combines positional non-local blocks with an Enhanced FPN architecture, utilizing coordinate maps for effective feature extraction in lane detection. Ablation studies (shown in Appendix \ref{app_A}) indicate that without Focusing Sampling, standard non-local blocks perform similarly to positional ones, suggesting lane coordinates need not necessarily be injected at non-local locations. Thus, FENetV2 aims to boost network efficiency by incorporating standard non-local blocks and exploring Focusing Enhanced FPN (FEFPN) and Directional IoU. This version seeks to maintain effectiveness while improving computational efficiency.

\textbf{FEFPN Structure.} As shown in Fig.~\ref{framework_fig}, most architecture configurations exhibit substantial similarities with PEFPN, with the distinction that standard non-local blocks are used during the construction of internal layers (Inter0, Inter1, Inter2).

\subsection{Lane Directional Intersection over Union (D-IoU) Module}
\begin{figure}[htbp]
\centering
\includegraphics[width=1\linewidth]{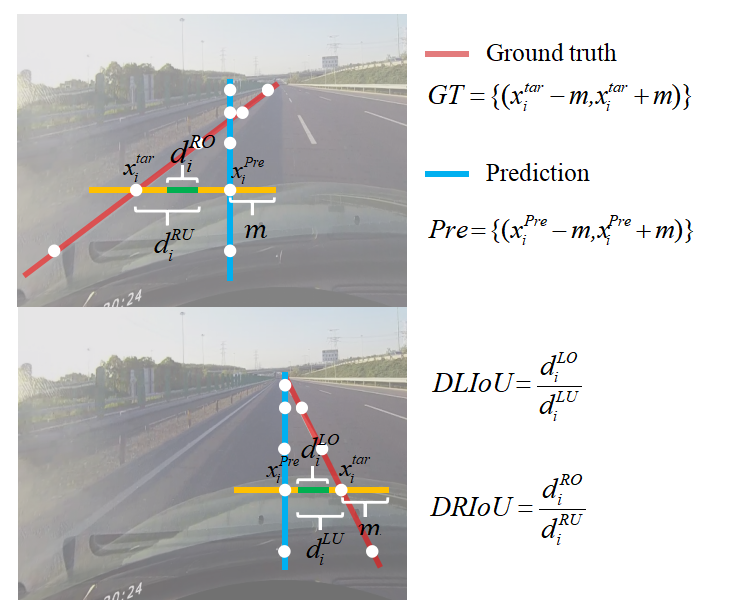}
\caption{\textbf{D-IoU Loss.} D-IoU loss assesses lane prediction accuracy by comparing predicted lanes to ground truth, using the IoU of extended lane segments from sampled points along the lane. This directional, localized loss metric accurately reflects predict precision across the entire lane geometry.}
\label{D-IOU_fig}
\end{figure}

\textbf{Motivation.} LineIoU only considers distance without directional relation \cite{4}. Our D-IoU module, Fig.~\ref{D-IOU_fig}, ascertains directional discrepancies to improve accuracy.

\textbf{Formulation.} As depicted, D-IoU comprises Position-IoU ($P_{IoU}$), Direction Left IoU ($DL_{IoU}$) and Direction Right IoU ($DR_{IoU}$). $P_{IoU}$ is the LineIOU module in the CLRNet. $DL_{IoU}$ and $DR_{IoU}$ represent the Distance-IoU to the Left and Right of the ground truth points, respectively. $DLIoU_i$ is:

{\small  \begin{align}
DLIoU_{i} & = \frac{d_{i}^{LO} }{d_{i}^{LU} } & = \frac{x_{i}^{tar}-max(x_{i}^{pre}-m,x_{i}^{tar}-m)}{m} 
\end{align}}

$m$ designates the pixel expansion amount for each point. After expanding by m pixels, $x_{i}^{pre}-m$ and $x_{i}^{pre}+m$ give the predicted point's left and right coordinates, respectively. The $DRIoU_{i}$ is a mirror mapping of the $DLIoU_{i}$, with similar definitions and will not be repeated.

D-IoU combines these using coefficients $\alpha$, $\beta$, $\gamma$:

{\small \begin{align}
D-IoU = \alpha (1-P_{IoU} )+\beta (1-DL_{IoU} )+\gamma (1-DR_{IoU})
\end{align}}

This provides distance and directional accuracy for precise lane alignment.

\subsection{Training and Inference Details}
\textbf{Training Loss.} The total training loss in FENetV1 is a weighted combination of several loss components:

{\small \begin{align}
L_{totalv1} = \omega_{Piou}L_{Piou} + \omega_{cls}L_{cls} + \omega_{xytl}L_{xytl} + \omega_{se}L_{se}
\end{align}}

$L_{Piou}$ is the PIoU Loss, weighted by $\omega_{Piou}$, which aligns predictions with ground truth position. $L_{cls}$ is the focal loss for classification. $L_{xytl}$ represents regression loss for predicting start point, angle, and lane length. $L_{se}$ is the semantic segmentation loss. The weights $\omega$ balance the contributions of each component.
The training loss in FENetV2 is:
{\small \begin{align}
L_{totalv2} = \omega_{Diou}L_{Diou} + \omega_{cls}L_{cls} + \omega_{xytl}L_{xytl} + \omega_{se}L_{se}
\end{align}}
Where $L_{Diou}$ is the D-IoU Loss, weighted by $\omega_{Diou}$, which aligns predictions with ground truth position and direction.


\begin{figure}[htbp]
\centering
\includegraphics[width=7cm]{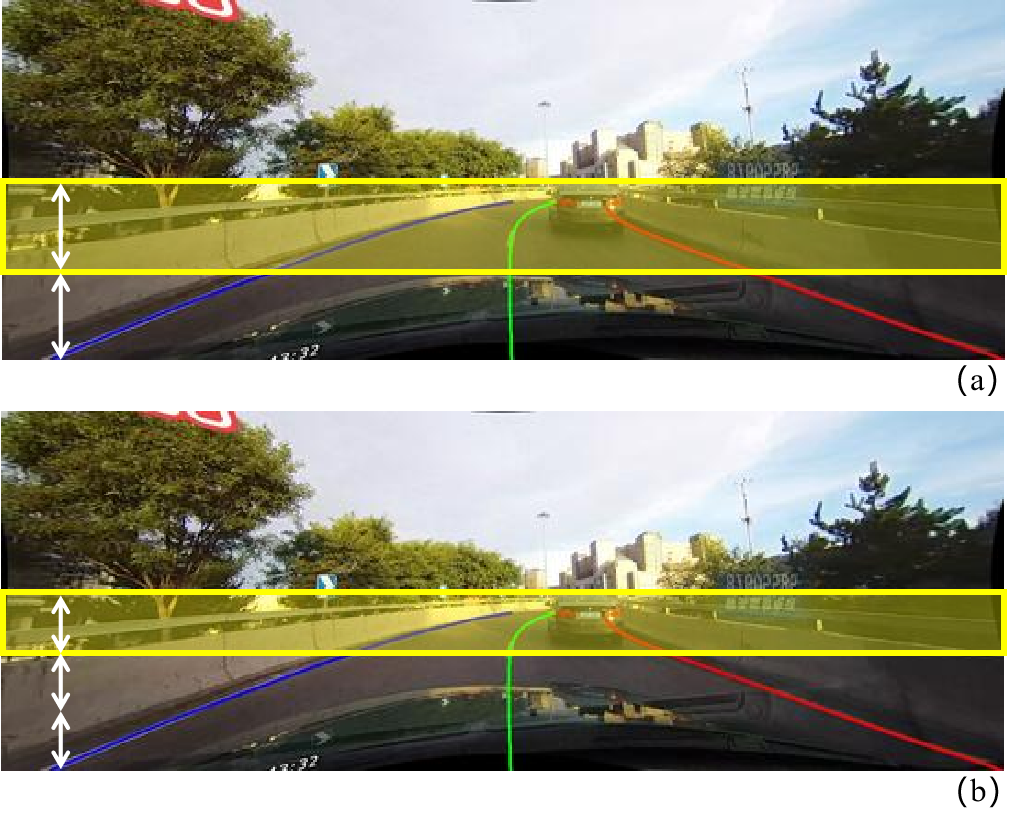}
\caption{The proposed \textbf{Partial Field of View} metric enhances evaluation by subdividing the lower image half into distal fraction views after preprocessing cropping. Assessing model precision on the (a) top 1/2 field and (b) top 1/3 field aligns with driving gaze ahead needs. This practical metric gauges lane detection aptitude beyond existing methods.}
\label{half_fig}
\end{figure}

\subsection{Augmenting Evaluation via Partial Field of View}


Current metrics for evaluating lane detection models, like pixel-wise accuracy and standard mAP, fail to reflect the real-world driving needs, treating all lane marking pixels equally despite varying safety impacts. Notably, experienced drivers focus on distant road sections, vital for predicting path geometry and steering adjustments. To better align evaluations with actual driving requirements, we suggest using a Partial Field of View metric. This method divides the lower half of an image into segments after cropping out the top irrelevant area. As shown in Fig.~\ref{half_fig}, this metric evaluates accuracy in critical distant areas (top half and top third of the field), closely matching where drivers typically look, thus providing a more accurate reflection of lane detection performance through the perspective of human driving behavior.

\section{Experiment}
\subsection{Datasets}
In this experiment, we employ the most commonly used datasets in lane detection: CULane \cite{7} and LLAMAS \cite{8}. CULane is currently one of the most challenging large-scale datasets for lane detection. LLAMAS is a comprehensive lane detection dataset comprising over 100,000 images. Lane markings within LLAMAS are auto-annotated employing high-precision maps.

\begin{table}[]
\centering
\setlength{\tabcolsep}{0.6mm}{
\begin{tabular}{llllll}
\Xhline{1px}
\textbf{Method} & \textbf{Backbone} & \textbf{mF1} &  \textbf{F1@50} &  \textbf{F1@75} & \textbf{GFlops} \\ \hline
UFLDV2       & ResNet18  & \textbf{-}     & 75    & \textbf{-}     & \textbf{-} \\
UFLDV2       & ResNet34  & -              & 76    & -              & -          \\
FOLOLane     & ERFNet    & -              & 78.8  & -              & -          \\
LaneAF       & ERFNet    & 48.6           & 75.63 & 54.53          & 22.2       \\
LaneAF       & DLA34     & 50.42          & 77.41 & 56.79          & 23.6       \\
LaneATT      & ResNet18  & 47.35          & 75.13 & 51.29          & 9.3        \\
LaneATT      & ResNet34  & 49.57          & 76.68 & 54.34          & 18         \\
LaneATT      & ResNet122 & 51.48          & 77.02 & 57.5           & 70.5       \\
GANet-S      & ResNet18  & -              & 78.79 & -              & -          \\
GANet-M      & ResNet34  & -              & 79.39 & -              & -          \\
GANet-L      & ResNet101 & -              & 79.63 & -              & -          \\ \hline
CLRNet       & ResNet18  & 55.23          & 79.58 & 62.21          & 11.9       \\
CLRNet       & ResNet34  & 55.14          & 79.73 & 62.11          & 21.5       \\
CLRNet       & ResNet101 & 55.55          & 80.13 & 62.96          & 42.9       \\
CLRNet       & DLA34     & 55.64          & \textbf{80.47} & 62.78          & 18.5       \\ \hline
\textbf{FENetV1(Ours)}  & DLA34     & \textbf{56.27} & 80.15 & \textbf{63.66} & 19.05     \\
\textbf{FENetV2(Ours)} & DLA34     & 56.17 & 80.19 & 63.50 & 18.85     \\ \Xhline{1px}
\end{tabular}%
}
\caption{FENet frameworks deliver top CULane performance with efficient GFlops usage. FENetV1 achieves the highest mF1 and F1@75 scores, outperforming all methods, including CLRNet, in lane recognition. FENetV2 performs comparably to V1, but is recommended for practical use, with its advantages detailed in subsequent sections.}
\label{tab:my-table}
\end{table}

\subsection{Implementation Details}
Primarily, DLA34 is used as the backbone network for pre-training in this study \cite{22}. Under the DLA34 backbone network, the CULane dataset is set to iterate for 15 epochs while LLAMAS is set to 20. The initial learning rate is set to 1e-3, the optimizer is AdamW, and the power is set at 0.9. The number of lane priors (N) is 72, and the number of sampling points is 36. In the expansion pixels (m) for $P_{IoU}$, $DL_{IoU}$  and $DR_{IoU}$ are all set to 15. The coefficient of assigning cost is set as $\omega _{Diou}$ = 1.

\subsection{Evaluation Metrics}
\textbf{F1 and mF1.} In the F1 test, the IoU is calculated between the prediction and the ground truth, and when the IoU threshold is greater than 0.5, it is considered a True Positive. The F1 is defined as:

{\small \begin{align}
F_{1} & = \frac{2\times Precision \times Recall}{Precision \times Recall} 
\end{align}}

About the COCO detection metric, we primarily continue to use the mF1 metric as the following metric in CLRNet, but to measure not only the overall performance but also the \textbf{scenes' performance}, providing a more precise analysis of prediction accuracy and lane feature capture. The mF1 is defined as:

{\small \begin{align}
mF_{1} & = (F1@50+F1@55+...+F1@95)/10
\end{align}}

where F1@50, F1@55, ..., F1@95 correspond to the F1 test values when the IoU threshold is 0.5, 0.55, ..., and 0.95, respectively. This is a more precise measurement in different scenes and has a significant effect on optimization iteration.

\begin{table*}[]
\centering
\resizebox{\textwidth}{!}{%
\begin{tabular}{llllllllllll}
\Xhline{1px}
\textbf{Field of View} &
  \textbf{Backbone} &
  \textbf{mF1} &
  \textbf{Normal} &
  \textbf{Crowded} &
  \textbf{Dazzle} &
  \textbf{Shadow} &
  \textbf{No line} &
  \textbf{Arrow} &
  \textbf{Curve} &
  \textbf{Cross} &
  \textbf{Night} \\ \hline
  
\textbf{Whole View of Lane} &
   &
   &
   &
   &
   &
   &
   &
   &
   &
   &
   \\
CLRNet &
  DLA34 &
  55.64 &
  68.72 &
  53.81 &
  47.4 &
  53.24 &
  35.28 &
  65.56 &
  40.62 &
  1154 &
  49.59 \\
FENetV1 (Ours) &
  DLA34 &
  \textbf{56.27} &
  68.7 &
  \textbf{55.12} &
  \textbf{48.16} &
  52.77 &
  \textbf{35.32} &
  {65.57} &
  {42.11} &
  \textbf{1147} &
  {50.51} \\
FENetV2 (Ours) &
  DLA34 &
  {56.17} &
  \textbf{69.19} &
  {54.38} &
  {47.67} &
  \textbf{53.39} &
  35.15 &
  \textbf{66.03} &
  \textbf{43.29*} &
  1206 &
  \textbf{50.55} \\ \hline
\textbf{Top 1/2 View of Lane} &
   &
   &
   &
   &
   &
   &
   &
   &
   &
   &
   \\ 
CLRNet &
  DLA34 &
  60.17 &
  73.83 &
  57.55 &
  53.24 &
  57.69 &
  \textbf{38.8} &
  69.81 &
  38.88 &
  1155 &
  55.3 \\
FENetV1 (Ours) &
  DLA34 &
  \textbf{61.08} &
  {74.11} &
  \textbf{59.09} &
  \textbf{53.92} &
  {58.46} &
  38.48 &
  {69.98} &
  {41.92} &
  \textbf{1147} &
  {56.69} \\
FENetV2 (Ours) &
  DLA34 &
  {60.97} &
  \textbf{74.59} &
  {58.21} &
  {53.75} &
  \textbf{58.5} &
  38.59 &
  \textbf{70.33} &
  \textbf{44.54*} &
  1206 &
  \textbf{56.73} \\ \hline
\textbf{Top 1/3 View of Lane} &
   &
   &
   &
   &
   &
   &
   &
   &
   &
   &
   \\ 
CLRNet &
  DLA34 &
  58.53 &
  72.46 &
  55.84 &
  52.84 &
  53.89 &
  38.48 &
  67.76 &
  31.1 &
  1155 &
  53.13 \\
FENetV1 (Ours) &
  DLA34 &
  \textbf{59.71} &
  73.06 &
  \textbf{57.45} &
  53.86 &
  \textbf{55.29} &
  38.28 &
  67.99 &
  34.61 &
  \textbf{1147} &
  55.04 \\
FENetV2 (Ours) &
  DLA34 &
  59.53 &
  \textbf{73.5} &
  56.43 &
  \textbf{54.03} &
  54.53 &
  \textbf{38.49} &
  \textbf{68.07} &
  \textbf{37.11*} &
  1206 &
  \textbf{55.05} \\ \Xhline{1px}
\end{tabular}%
}
\caption{\textbf{Performance by Field of View.} Compared to optimal CLRNet, FENet demonstrates mF1 gains over entire, the top 1/2, and top 1/3 views. Particularly significant curved and distant lane improvements manifest, with FENetV2 curve detection higher by 2.67, 5.66, and 6.01 on those views respectively. This highlights FENetV2’s advantages on challenging curves versus prior works. Overall, FENetV2 surpasses V1 on regression precision due to its D-IOU loss and FEFPN enabling optimized boundary localization. However, FENetV1 exhibits greater competency on lane recognition itself rather than regression due to its PEFPN inferring spatial layouts effectively.}
\label{tab:table3}
\end{table*}

\begin{table}[]
\setlength{\tabcolsep}{1.3mm}{
\begin{tabular}{c
>{}c ccc}
\Xhline{1px}
\multicolumn{1}{c}{\multirow{2}{*}{\textbf{Method}}} & \multicolumn{1}{c}{\multirow{2}{*}{\textbf{Backbone}}} & \multicolumn{3}{c}{\textbf{Valid}}                                                       \\ \cline{3-5} 
\multicolumn{1}{c}{}                        & \multicolumn{1}{c}{}                          & \multicolumn{1}{c}{\textbf{mF1}} & \multicolumn{1}{c}{\textbf{F1@50}} & \multicolumn{1}{c}{\textbf{F1@75}} \\ \hline
PolyLaneNet                                 & EfficientnetB0                                                                     & 48.82                                                         & { 90.2}           & { 45.4}           \\
LaneATT                                     & { ResNet-18}                                                   & 69.22                                                         & 94.64                                                         & 82.36                                                         \\
LaneATT                                     & { ResNet-34}                                                   & 69.63                                                         & 94.96                                                         & 82.79                                                         \\
LaneATT                                     & { ResNet-122}                                                  & 70.8                                                          & 95.17                                                         & 84.01                                                         \\
LaneAF                                      & { DLA-34}                                                      & 69.31                                                         & 96.9                                                          & 84.71                                                         \\

{ CLRNet}               & { DLA-34}                                                      & 71.21                                                         & \textbf{97.16}                                                & 85.33                                                         \\ \hline
{ \textbf{FENetV2(Ours)}} & { DLA-34}                                                      & { \textbf{71.85}} & { 96.97}          & { \textbf{85.63}} \\ \Xhline{1px}
\end{tabular}}
\caption{FENetV2 establishes new \textbf{state-of-the-art LLAMAS} results, proving advanced lane detection aptitude. Our approach sets the top mF1 using the DLA-34 backbone while exceeding all methods on the vital long-range F1@75.}
\label{tab:table2}
\end{table}

\subsection{Comparison with State-of-the-Art Results}

\textbf{Performance on CULane.} Our proposed FENet achieves state-of-the-art results on the CULane benchmark, surpassing prior methods. As shown in Table 1, FENetV1 obtains an F1@75 score of \textbf{63.63} and an mF1 of \textbf{56.27}, \textbf{exceeding} CLRNet's F1@75 by \textbf{0.7} and mF1 by \textbf{0.63}. FENetV2 obtains an F1@75 score of \textbf{63.50} and an mF1 of \textbf{56.17}, this demonstrates FENetV1's and FENetV2's precision for lane detection, particularly at stricter evaluation thresholds.


\textbf{Performance by Field of View.} As shown in Table \ref{tab:table3}, FENetV2 outperforms the CLRNet model in detecting curved and distant lanes across all, the top half, and the top third fields of view. This improvement is crucial in driving scenarios where bends are primarily visible from afar, enhancing safe manoeuvring by better detecting distant curves. Specifically, FENetV2 achieves significantly higher mF1 scores of \textbf{2.67}, \textbf{5.66}, and \textbf{6.01} in these * data, underscoring its ability to accurately identify challenging distant and curved lanes. This performance is attributed to its D-IoU loss function and FEFPN module, with detailed examples provided in Fig.~\ref{fig8} in Appendix \ref{app_A}.


Although FENetV1 scores slightly higher in overall mF1 metrics, FENetV2 is more effective and reliable for real-world autonomous driving due to its focus on distant lane boundary regression. This capability is crucial for accurate lane localization needed for quick vehicle control at high speeds. While FENetV1 excels in general lane detection, it lacks precision in boundary localization. In essence, FENetV2 stands out in practical lane detection performance by specializing in distant lane regression, making it preferable for autonomous navigation where immediate responses are critical. We recommend FENetV2 over other models, including FENetV1, for its superior real-world applicability.

\textbf{Performance on LLAMAS.} As Table \ref{tab:table2} shows, the innovative FENetV2 structure proposed in this study achieves a new technical level on LLAMAS, with an F1@75 score of \textbf{85.63} and an mF1 score of \textbf{71.85}. These major scoring parameter results are higher than those of CLRNet, with the F1@75 score being \textbf{0.3 higher} than of CLRNet and the mF1 score being \textbf{0.64 higher} than that of CLRNet. This also indicates that our method makes improvements in lane detection accuracy.


\section{Conclusion}
Inspired by human driving focus, this research pioneers Focusing on Enhanced networks, sampling strategies, optimized loss calculations, and refined evaluation metrics targeting lane detection challenges for autonomous navigation. Experiments demonstrate emphasizing critical distant geometric details, unlike existing uniform approaches, significantly improves not only benchmark accuracy but also practical curved/distant lane recognition essential for safety. Advancements derive from isolating perspective-aware contexts mimicking adept driver vision patterns. Limitations provide opportunities including refining attention regions, exploring enriched coordinate representations, collecting actual driving data for analysis, and reconciling dual frameworks exploiting complementary strengths - furthering real-world breakthroughs. With human-mimicking visual perception and comprehension as a guide, the lane detection frontier can rapidly advance toward enabling reliable autonomous vehicle control.

\bibliographystyle{IEEEbib}
\bibliography{icme2023template}

\appendix
\clearpage

\section{Appendix}
\subsection{Ablation Studies}
\label{app_A}

To validate the contributions and roles of each component of the entire FENetV1 and FENetV2 model in the overall experiment, we separately test each innovative scheme on the CULane dataset to display their corresponding performance.

\subsection{Overall Ablation Study}

Through ablation studies, we analyse our proposed FENetV1 architecture in Appendix \ref{app_A} Table \ref{a1} to analyse the contributions of the PEFPN, Focusing Sampling, and D-IoU modules. The baseline FENetV1 achieved an mF1 of 55.64. With the addition of the PEFPN and Focusing Sampling modules, the mF1 increased to 56.27, demonstrating their benefits in providing positional-aware features and emphasising hard examples. However, surprisingly, incorporating D-IoU on top of PEFPN and Focusing Sampling resulted in a slight decrease in mF1 to 56.04. We hypothesise that this counterintuitive result is due to some functional overlap between the position-aware capabilities of PEFPN and the directional encoding of D-IoU. Specifically, the positional non-local block within PEFPN already integrates coordinate information into the multi-scale features. Thus, the directional modelling of D-IoU becomes somewhat redundant and interferes with the positional encodings of PEFPN. This integration of position and direction within PEFPN may explain why the subsequent addition of D-IoU leads to detrimental effects due to duplicate functionality.

Further ablation experiments are conducted on the contribution of each component of our proposed FENetV2 model in Appendix Table \ref{a2}. The baseline architecture achieves an mF1 of 55.64. Adding the FEFPN module improves mF1 to 56.11 by providing richer multi-scale features. Incorporating Focusing Sampling further boosts mF1 to 56.15 by emphasising hard far-end examples. Finally, replacing the IoU loss with the D-IoU loss increases mF1 to 56.17 by encoding orientation cues. The steady improvements validate the benefits of FEFPN for contextual features, Focusing Sampling for handling distant lanes, and D-IoU for differentiating directionality.

\begin{table*}[]
\centering
\fontsize{10}{12}\selectfont
\resizebox{\textwidth}{!}{%
\begin{tabular}{ccccccccc}
\Xhline{1px}
\textbf{PEFPN} & \textbf{Focusing Sampling} & \textbf{D-IoU} & \textbf{mF1}   & { \textbf{F1@50}} & { \textbf{F1@60}} & { \textbf{F1@70}} & { \textbf{F1@80}} & { \textbf{F1@90}} \\ \hline
               &                          & \textbf{}      & 55.64          & \textbf{80.47}                & 76.15                & 68.87                & 53.95                & 20.42                \\ \hline
$\surd $               &                          &                & 56.11          & 80.04                & 76.32                & 69.32                & 54.71                & 21.46                \\ \hline
$\surd $               & $\surd $                         &                & \textbf{56.27} & 80.16       & \textbf{76.54}       & \textbf{69.54}       & \textbf{55.02}       & 21.53                \\ \hline
$\surd $               & $\surd $                         & $\surd $             & 56.04 & 80.04       & 76.24       & 69.26                & 54.66                & \textbf{21.6}        \\ \Xhline{1px}
\end{tabular}
}
\caption{The effects of each module in the FENetV1 method. Results based on CULane.}
\label{a1}
\end{table*}

\begin{table*}[]
\centering
\resizebox{\textwidth}{!}{%
\begin{tabular}{ccccccccc}
\Xhline{1px}
\textbf{FEFPN}       & \textbf{Focusing Sampling} & \textbf{D-IoU} & \textbf{mF1}   & { \textbf{F1@50}} & { \textbf{F1@60}} & { \textbf{F1@70}} & { \textbf{F1@80}} & { \textbf{F1@90}} \\ \hline
\multicolumn{1}{l}{} &                          & \textbf{}      & 55.64          & \textbf{80.47}                & 76.15                & 68.87                & 53.95                & 20.42                \\ \hline
$\surd $                    &                          &                & 56.11          & 80.24               & 76.35                & 69.39                & 54.82                & 21.55                \\ \hline
$\surd $                    & $\surd $                        &                & 56.15          & 80.04                & 76.35                & \textbf{69.49}       & \textbf{55.14}       & 21.44                \\ \hline
$\surd $                    & $\surd $                        & $\surd $              & \textbf{56.17} & 80.19      & \textbf{76.36}       & 69.2                 & 54.93                & \textbf{21.79}       \\ \Xhline{1px}
\end{tabular}
}
\caption{The effects of each module in the FENetV2 method. Results based on CULane.}
\label{a2}
\end{table*}

In summary, the ablation study of FENetV1 reveals that Enhanced FPN with position-aware representations through PEFPN and emphasising the complementary strengths of hard samples via Focusing Sampling improved mf1 to 56.27. However, the D-IoU conceived with the ideology of positional non-local blocks may exhibit redundancy with the positional modelling already encoded within PEFPN, culminating in a slight degradation in mF1 performance. In contrast, the stable mF1 enhancement conferred by each constituent in the FENetV2 ablation experiments verifies their efficacy in assimilating rich multi-scale features, accentuating challenging regions, and embedding directional clues within our FENetV2 framework. Despite a minor decline in mF1 compared to FENetV1, superior scene-fitting accuracy is demonstrated in the manuscript. This analysis proffers constructive perceptions into the architectural trade-offs between positional non-local blocks and D-IoU formulations.

\subsection{Ablation Study of Focusing Sampling}

To further analyse the efficacy of our proposed Focusing Sampling technique, we conduct ablation studies comparing models with and without Focusing Sampling in Appendix Table \ref{a3}. Using uniform sampling as the baseline, the model achieved an mF1 of 55.64. Replacing this with Focusing Sampling provided a slight boost to 55.78, indicating its benefits for emphasising challenging examples. The advantages of Focusing Sampling become more pronounced when coupled with our feature-enhanced FPN modules. With uniform sampling, the FEFPN and PEFPN models obtain similar mF1 scores of 56.11. However, the addition of Focusing Sampling improves their mF1 to 56.15 and 56.27 respectively. This demonstrates that Focusing Sampling better utilises the rich lane features provided by FEFPN and PEFPN, by concentrating training on the most difficult far-end regions. Notably, FEFPN and PEFPN achieve the same mF1 without Focusing Sampling. This suggests that with uniform focus, neither FPN variant could fully leverage their learned features. However, by centralizing attention, PEFPN integrated positional information to boost performance above FEFPN. Overall, these ablation studies validate that Focusing Sampling effectively complements advanced FPN modules by enabling concentrated learning on hard examples. The gains are amplified when combined with FPN designs that encode multi-scale semantics and spatial coordinates.

\begin{table*}[]
\centering
\fontsize{12}{14}\selectfont
\resizebox{\textwidth}{!}{%
\begin{tabular}{cccccccll}
\Xhline{1px}
\textbf{Sampling Settings} & \textbf{mF1}   & { \textbf{F1@50}} & { \textbf{F1@60}} & { \textbf{F1@70}} & { \textbf{F1@80}} & { \textbf{F1@90}} &  &  \\ \hline
Uniform sampling             & 55.64          & \textbf{80.47}       & 76.15                & 68.87                & 53.95                & 20.42                &  &  \\ \hline
Uniform sampling + FEFPN     & 56.11          & 80.24                & 76.35                & 69.39                & 54.82                & 21.55                &  &  \\ \hline
Uniform sampling + PEFPN     & 56.11          & 80.04                & 76.32                & 69.32                & 54.71                & 21.46                &  &  \\ \hline
Focusing Sampling          & 55.78          & 79.93                & 76.03                & 68.72                & 54.24                & \textbf{21.56}       &  &  \\ \hline
Focusing Sampling + FEFPN  & 56.15          & 80.04                & 76.35                & 69.49                & \textbf{55.14}       & 21.44                &  &  \\ \hline
Focusing Sampling + PEFPN  & \textbf{56.27} & 80.16       & \textbf{76.54}       & \textbf{69.54}       & 55.02                & 21.53                &  &  \\ \Xhline{1px}
\end{tabular}
}
\caption{The ablation research for Focusing Sampling in  PEFPN(FENetV1) and FEFPN(FENetV2) methods. Results based on CULane.}
\label{a3}
\end{table*}

\begin{figure*}[htbp]
\centering
\includegraphics[width=17cm]{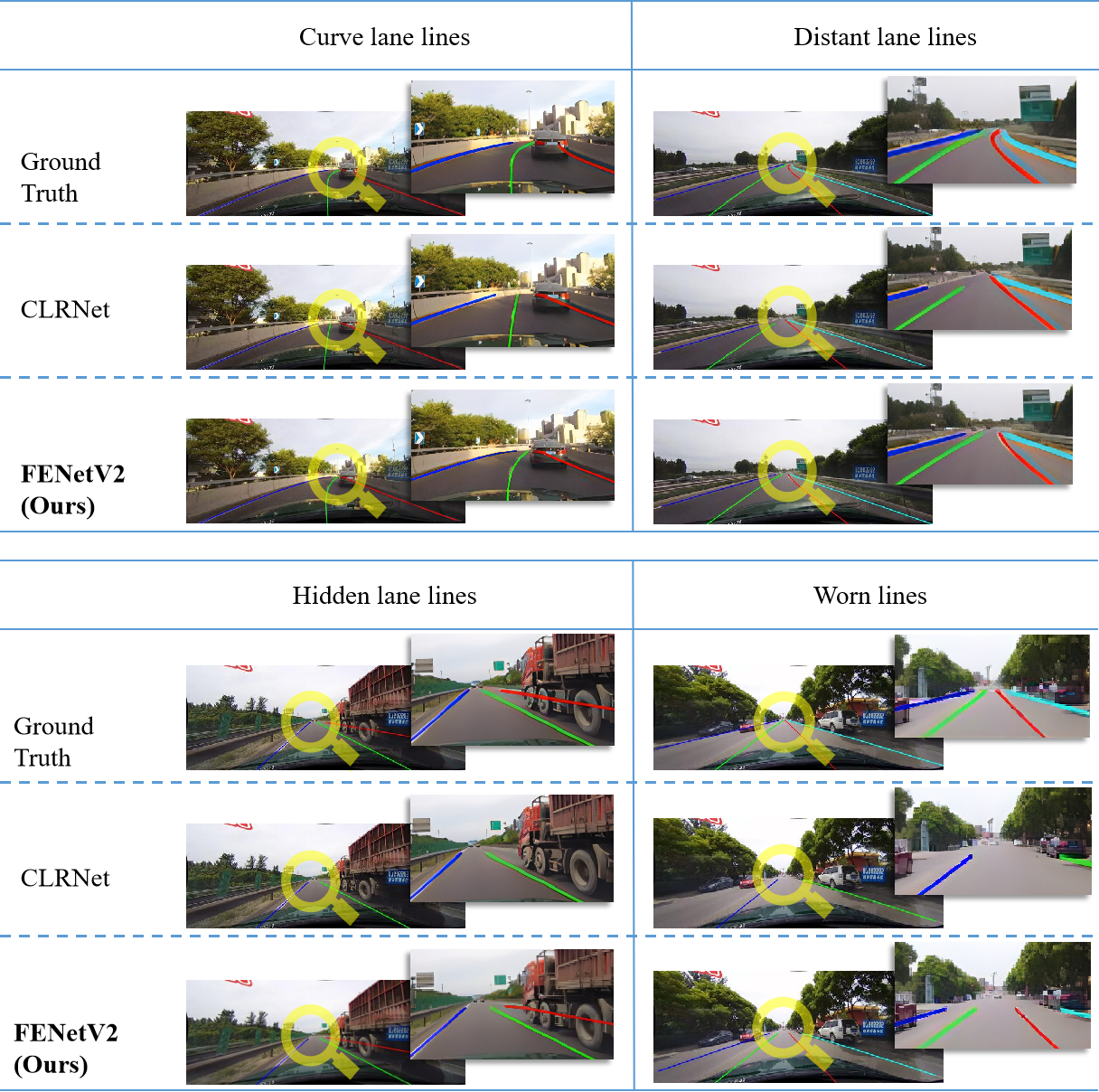}
\caption{Comparison of the detection effect of curve, distant, hidden and worn lane lines on some difficult samples with ground truth. The upper right corner of each image results from a 4x pixel magnification of the human eye focus position for the front ahead.
}
\label{fig8}
\end{figure*}

\clearpage

\begin{figure*}[htbp]
\centering
\includegraphics[width=9cm]{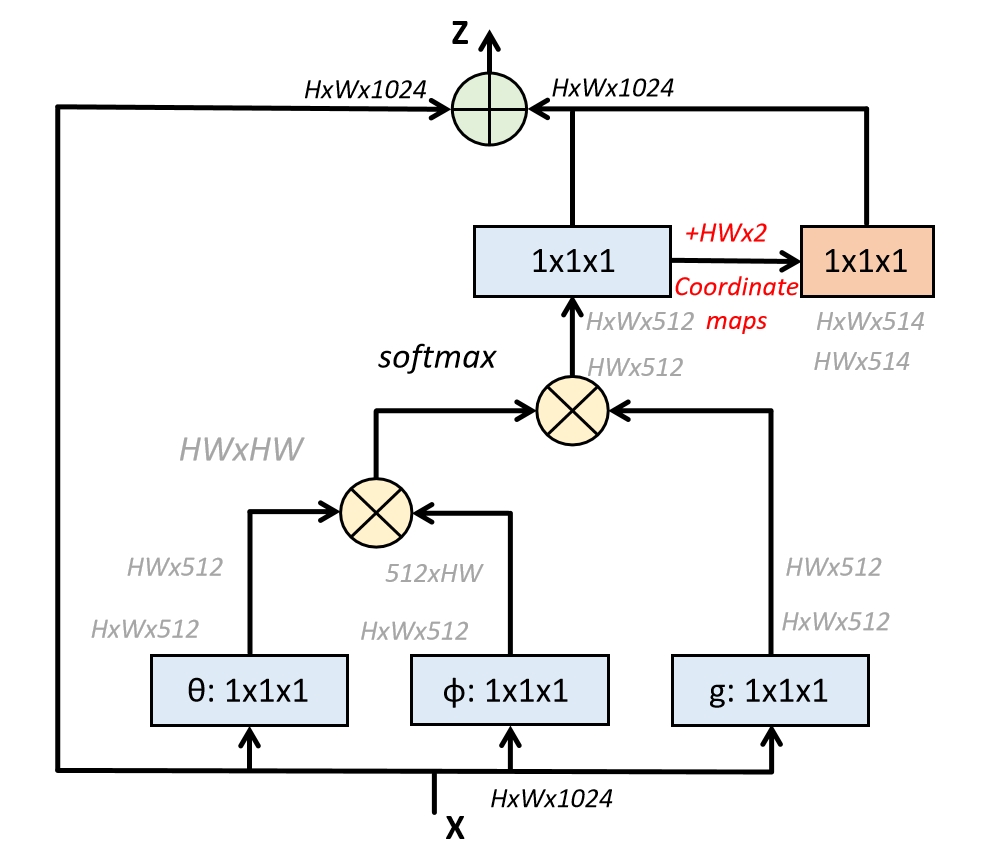}
\caption{Illustration of positional non-local blocks. The extra coordinate maps are fused at the rear of the original non-local block.}
\label{fig6}
\end{figure*}

\begin{figure*}[htbp]
\centering
\includegraphics[width=1\linewidth]{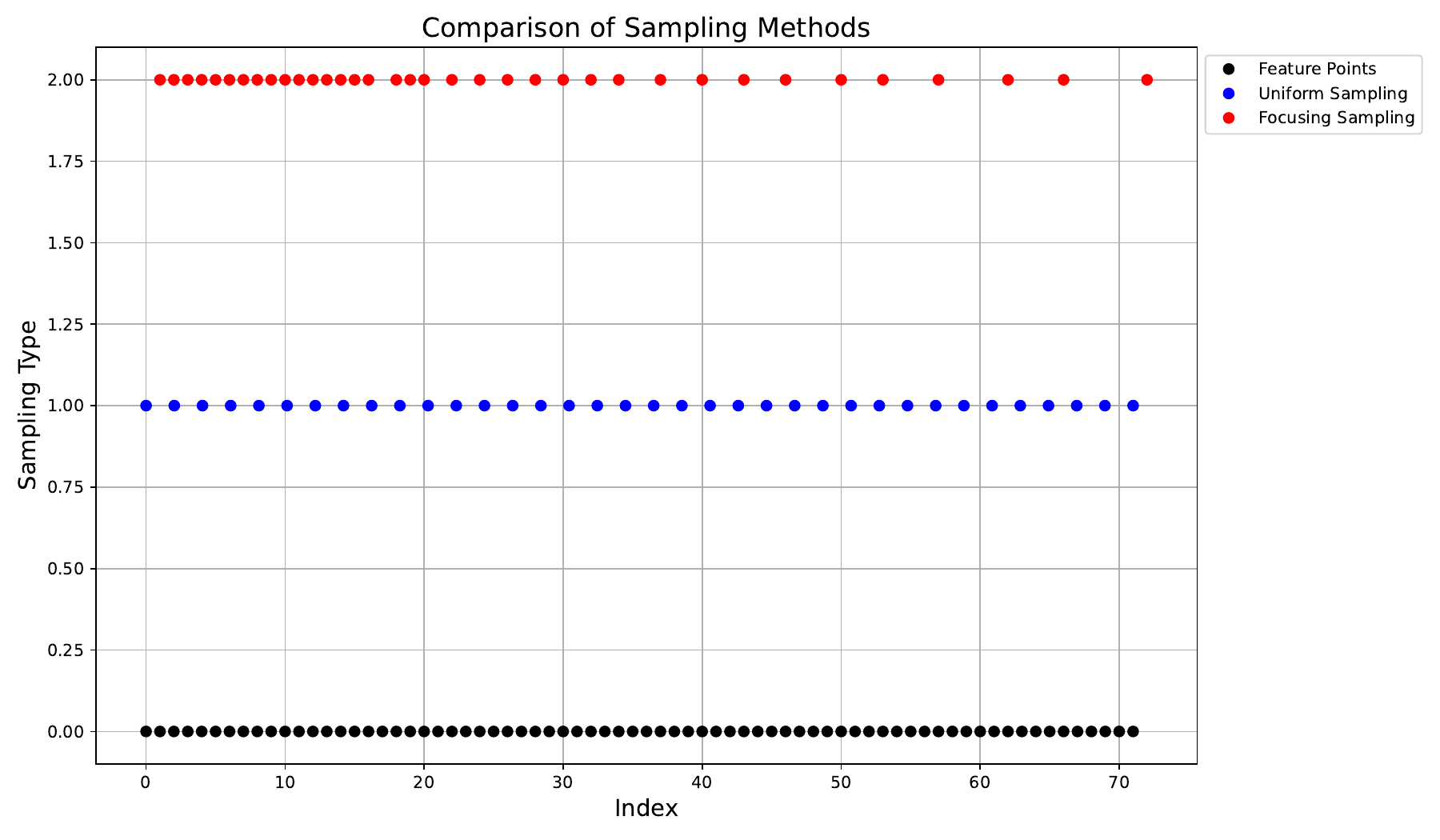}
\caption{Focusing Sampling and uniform sampling intuitive comparison. The black dots in the figure denote 72 feature points. The blue dots signify feature points selected via uniform sampling, while the red dots mark feature points chosen through Focusing Sampling. The feature point Focusing Sampling in this work proceeds from densely populated feature point regions in distant areas of the visual scene, progressing toward more sparsely distributed features in proximity to the observer. 
}
\label{fig7}
\end{figure*}

\end{document}